\documentclass{article} 
\usepackage[preprint]{colm2025_conference}

\usepackage{microtype}
\usepackage{hyperref}
\usepackage{url}
\usepackage{booktabs}
\usepackage{graphicx}
\usepackage{lineno}
\usepackage{xspace}
\usepackage{booktabs}
\usepackage{algorithm}
\usepackage{algpseudocode}
\usepackage{amsmath}
\usepackage{subcaption}
\usepackage{comment}

\definecolor{darkblue}{rgb}{0, 0, 0.5}
\hypersetup{colorlinks=true, citecolor=darkblue, linkcolor=darkblue, urlcolor=darkblue}

\newcommand{\dumb}{\textsc{dumb500}\xspace}

\newcommand{\method}{\textbf{\textsc{ThoughtTerminator}}\xspace}
\newcommand{\methodfancy}{\textbf{\textsc{\textcolor[rgb]{0,0,0}{T}\textcolor[rgb]{0.2,0,0}{h}\textcolor[rgb]{0.4,0,0}{o}\textcolor[rgb]{0.5,0,0}{u}\textcolor[rgb]{0.6,0,0}{g}\textcolor[rgb]{0.7,0,0}{h}\textcolor[rgb]{0.8,0,0}{t}\textcolor[rgb]{0.85,0,0}{T}\textcolor[rgb]{0.9,0,0}{e}\textcolor[rgb]{0.95,0,0}{r}\textcolor[rgb]{1,0,0}{m}\textcolor[rgb]{1,0,0}{i}\textcolor[rgb]{1,0,0}{n}\textcolor[rgb]{1,0,0}{a}\textcolor[rgb]{1,0,0}{t}\textcolor[rgb]{1,0,0}{o}\textcolor[rgb]{1,0,0}{r}}}\xspace}

\renewcommand{\cite}[1]{\citep{#1}}

\newif\ifcommentsoff

\title{\vspace{-3pt}\raisebox{-3pt}{\includegraphics[width=20pt]{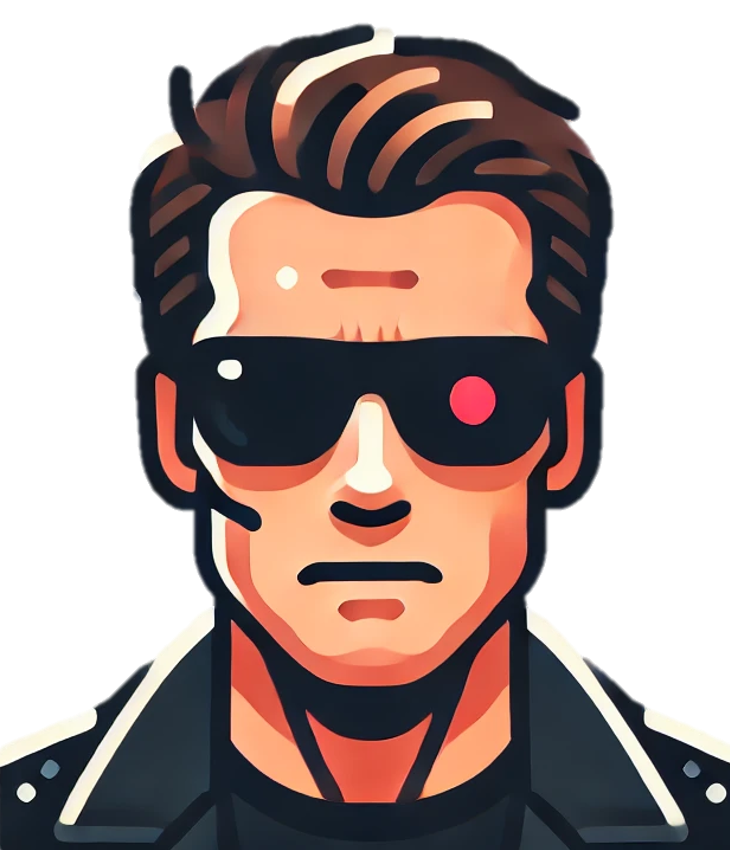}} \methodfancy: Benchmarking, Calibrating, and Mitigating Overthinking in Reasoning Models}

\author{Xiao Pu\thanks{ Co-first contributions.}\;\;\quad Michael Saxon$^*$\quad  Wenyue Hua\quad William Yang Wang\\
University of California, Santa Barbara\\
Contact: \texttt{xiao\_pu@ucsb.edu}, \texttt{saxon@ucsb.edu}
}

\begin{document}

\ifcolmsubmission
\linenumbers
\fi

\maketitle

\begin{abstract}
Reasoning models have demonstrated impressive performance on difficult tasks that traditional language models struggle at. However, many are plagued with the problem of overthinking---generating large amounts of unnecessary tokens which don't improve accuracy on a question. We introduce approximate measures of problem-level difficulty and demonstrate that a clear relationship between problem difficulty and optimal token spend exists, and evaluate how well calibrated a variety of reasoning models are in terms of efficiently allocating the optimal token count. We find that in general, reasoning models are poorly calibrated, particularly on easy problems. To evaluate calibration on easy questions we introduce \dumb, a dataset of extremely easy math, reasoning, code, and task problems, and jointly evaluate reasoning model on these simple examples and extremely difficult examples from existing frontier benchmarks on the same task domain. Finally, we introduce \method, a training-free black box decoding technique that significantly improves reasoning model calibration.
\end{abstract}

\section{Introduction}

Investment in improving the capabilities of language models has recently turned from data- and train-time-scaling to \textit{inference-scaling}, or training so-called \textit{reasoning models} to expend more runtime compute generating chains of thought \cite{Wei2022ChainOT}, debate \cite{Liang2023EncouragingDT}, and self-corrections \cite{Pan2024AutomaticallyCL} in order to more robustly and correctly answer queries \cite{Wu2024InferenceSL}.

On average, there is a direct relationship between amount of inference spend and performance on benchmarks of a variety of ``reasoning tasks'' \cite{jaech2024openai}.

Under the inference scaling paradigm, controlling costs is critical.
Unfortunately, open reasoning models such as DeepSeek r1 \cite{DeepSeekAI2025DeepSeekR1IR} and QwQ \cite{qwq32b} have demonstrated a tendency to expend unnecessary inference tokens after the answer has already could be generated, a problem referred to as \emph{overthinking} \cite{Chen2024DoNT}.

We need to precisely define overthinking in order to mitigate it.
\citet{Chen2024DoNT} define overthinking as the amount of times the model repeats the correct answer in its intermediate reasoning chain.
From this definition, they used supervised fine-tuning and direct preference optimization to train reasoning models to prefer to select the shortest answer.
Similar work applied knowledge distillation from non-reasoning models to blend their preference to answer concisely with the reasoning models' better performance \cite{Yang2025TowardsTS}.
However, both of these methods require retraining, a process that may be costly or have unintended consequences on performance.

Training-free methods which seek to manage overthinking include selective invocation of chain-of-thought on tasks where it has known benefit \cite{Sprague2024ToCO} early stopping of reasoning chains using probe-based confidence of final answer tokens \cite{fu2024efficiently}, or simply eliciting reasoning model-like behavior from non-reasoning models using continuing phrases like ``wait...'', which can be halted at any time \cite{Muennighoff2025s1ST}.
Limitations of these methods include requiring external knowledge of task type, white-box access to the base model, or the use of non-reasoning models for precise control \cite{Yu2025ThinkSN}.

In this work we seek to analyze the {difficulty calibration} of token spend in reasoning models. Starting from the supposition that more difficult problems require more thought, we first characterize this difficulty-cost relationship in a variety of open reasoning models across three reasoning datasets---MATH500 \cite{Lightman2023LetsVS}, GPQA \cite{Rein2023GPQAAG}, and ZebraLogic \cite{zebralogic2024}---allowing us to introduce a difficulty-calibrated measure of overthinking.

As these three existing datasets only allow us to assess overthinking in reasoning models on hard problems, we introduce \dumb, a dataset of `easy' queries to explore overthinking on easy inputs.

With the overthinking problem formally defined, we introduce \method, a \textit{training-free, black box decoding strategy to mitigate overthinking} using difficulty-calibrated conditioning.
We show that \method is a simple and effective way to control overthinking in reasoning models without requiring any access to gradients or training.

\begin{figure}[t!]
    \centering
    \includegraphics[width=\linewidth]{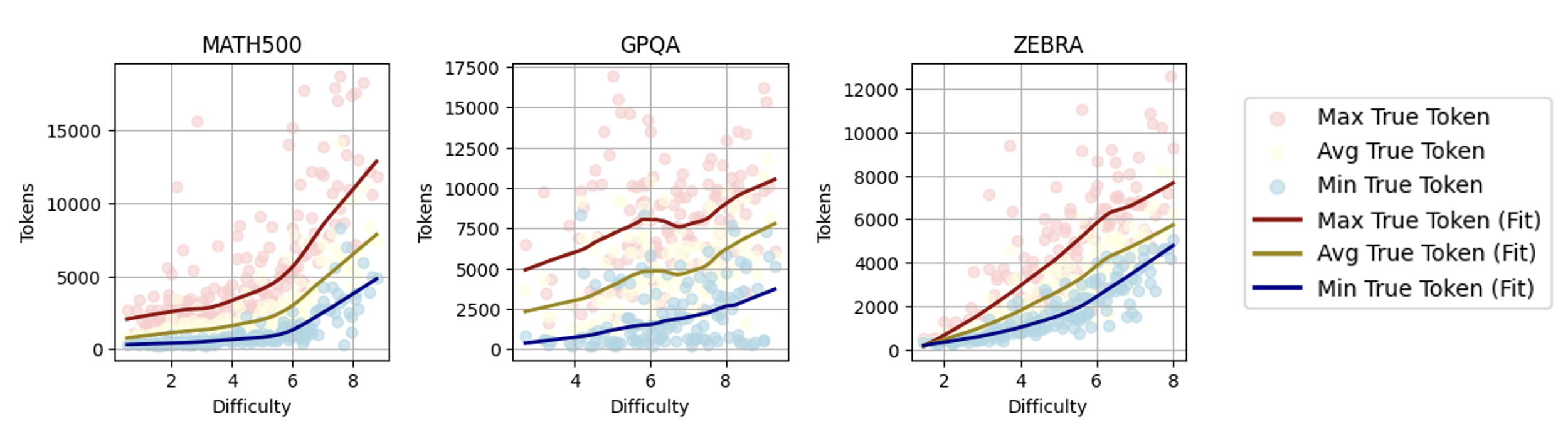}
    \caption{Question-level difficulty vs average token spend across models for three reasoning datasets. Difficulty scores are scaled by 10 and mapped to integers from 1 to 10 for readability.
    We observe a clear relationship between question difficulty and token spend distribution.}
    \label{fig:difficulty_scatter}
\end{figure}

\section{Difficulty Calibration in Reasoning Models}\label{sec:difficulty}

This work is concerned with how optimally reasoning models allocate \textit{token spend} $Sp$, or total number of tokens generated in a given sample to respond to an input.

Given that increased inference scale leads to higher performance across a variety of reasoning tasks, it is reasonable to hypothesize that the \textit{difficulty of a question correlates with optimal token spend.}
We characterize the difficulty $D$ of a given question $q$ pair for model $\mathcal{M}$ as the simple inaccuracy rate of that model over that pair over $n$ samples of that question $q$ and it's gold answer $a$.

\begin{equation}
    D_\mathcal{M}(q,a) = p(\hat{a} \sim
    \mathcal{M}(q) \neq a) 
    \approx \sum_n  \mathbb{1}(\mathcal{M}(q) \neq a) / n
\end{equation}

We can compute a multi-model difficulty estimate $\Bar{D}$ of $q$ as the expected difficulty $ \mathbb{E}[D(q,a)]$ over a class of models $\mathbf{M}$. While this definition is model-dependent, it captures an operational notion of difficulty that is both reproducible and relevant for analyzing inference efficiency under current LLMs. 

\begin{equation}
    \Bar{D}(q) = \mathbb{E}[D(q,a)] \approx \sum_{m \in \mathbf{M}} \sum_n \mathbb{1}(\mathcal{M}(q) \neq a) / |\mathbf{M}|n 
\end{equation}

Each answer $a_i$ incidentally sampled from $\mathcal{M}$ in response to question $q$ is associated with its own token spend $Sp_\mathcal{M}(a_i)$. Is there a relationship between the difficulty of each question and the token spend that naturally occurs?

We assess the difficulty $\Bar{D}$ and token spend $Sp_\mathcal{M}$ using reasoning and non-reasoning models from the DeepSeek \cite{DeepSeekAI2025DeepSeekR1IR}, Qwen \cite{qwen2.5,qwq32b}, Gemma \cite{Mesnard2024GemmaOM}, and LLaMa \cite{Dubey2024TheL3} families for all questions in the MATH500 \cite{Lightman2023LetsVS}, GPQA \cite{Rein2023GPQAAG}, and ZebraLogic \cite{zebralogic2024} datasets.

\begin{table}[t!]
    \centering
    \resizebox{.9\textwidth}{!}{
    \begin{tabular}{lrr}
    \toprule
    Model & Local overthinking $O_\textrm{env}$ $\downarrow$ & Global overthinking $O_g$ $\downarrow$ \\
    \midrule
    \multicolumn{3}{c}{Non-reasoning language models} \\
    \midrule
    \texttt{Qwen2-7B-Instruct}              & 291 & 219 \\
    \texttt{Llama-3.2-1B-Instruct}          & 542 & 354 \\
    \texttt{Llama-3.2-3B-Instruct}          & 708 & 473 \\
    \texttt{Llama-3.1-8B-Instruct}          & 1971 & 1755 \\
    \texttt{gemma-2-2b-it}                  & 148 & 152 \\
    \texttt{gemma-2-9b-it}                  & 131 & 161 \\
    \texttt{gemma-2-27b-it}                 & 178 & 187 \\
    \texttt{deepseek-llm-7b-chat}           & 155 & 90 \\
    \midrule
    \multicolumn{3}{c}{Reasoning language models} \\
    \midrule
    \texttt{QwQ-32B-Preview}                & 2923 & 3698 \\
    \texttt{QwQ-32B}                        & 13662 & 11248 \\
    \texttt{DeepSeek-R1-Distill-Qwen-1.5B}  & 5730 & 4262 \\
    \texttt{DeepSeek-R1-Distill-Llama-8B}   & 4232 & 5755 \\
    \texttt{DeepSeek-R1-Distill-Qwen-7B}    & 3881 & 4001 \\
    \bottomrule
    \end{tabular}}
    \caption{Local and global overthinking scores (rounded to integers).}
    \label{tab:overthinking-scores}
\end{table}

\autoref{fig:difficulty_scatter} contains scatter plots of $D_{\mathcal{M}}$ and $Sp(a)$ for each answer $a$ from \texttt{DeepSeek-R1-7B} for all three datasets. 
We observe that similar to the dataset \& model-wise relationships between performance and token spend documented in prior work \cite{Muennighoff2025s1ST}, there also exists a clear relationship between question-level difficulty and average token spend.

Additionally, we note \textit{considerable variance in the token spend between answer samples for each question.}
These reasoning models exhibit considerable inconsistency in their efficiency between samples.
This leads to two natural questions:
\begin{enumerate}
    \item How \textbf{well-calibrated} are reasoning models in consistently realizing their optimal token spend per-question?
    \item Is it possible to improve the calibration of reasoning models in their token spend?
\end{enumerate}

\subsection{Quantifying Overthinking}

We formalize \textbf{observational overthinking}, or the failure in consistency a reasoning model has at realizing the minimum possible token spend per question.

The \textit{observed minimum spend} of a question is the shortest reasoning chain of its full set of \textit{correct model-generated answers}. We measure observational overthinking in terms of the difference between a model's typical token spend and this observed minimum.
For questions sampled from dataset $\mathcal{D}$, the \textbf{global overthinking score} $O_g$ of a model is the mean difference between the length of each reasoning chain and the global observed minimum spend for each question.

\begin{equation}
    O_g(\mathcal{M}) = \sum_{q \in \mathcal{D}} \big( \mathbb{E} [Sp(a \sim \mathcal{M} | q)] - \min_{\mathcal{M}_i \in \mathbf{M}}(Sp(a \sim \mathcal{M}_i | q)) \big) / |\mathcal{D}|
\end{equation}

The \textbf{local envelope overthinking score} $O_\textrm{env}$ is the mean difference between the maximum and minimum spends for each question for each model.

\begin{equation}
    O_\textrm{env}(\mathcal{M}) = \sum_{q \in \mathcal{D}} \big( \max[Sp(a \sim \mathcal{M} | q)] - \min(Sp(a \sim \mathcal{M} | q)) \big) / |\mathcal{D}|
\end{equation}

\autoref{tab:overthinking-scores} presents the calibration scores for the full set of LLama, Qwen, Gemma, and DeepSeek models we evaluated on the three datasets. These calibration scores represent {expected quantities of tokens wasted}, as they are averages in excess of minimum spend values. \textbf{Lower is better.}
As expected, the reasoning models with propensity to overthink have considerably higher overthinking scores than the non-reasoning models.

One weakness of our overthinking evaluation so far is that we have very few questions that have a low difficulty but high overthinking tendency.
This is because reasoning models are evaluated mainly on challenging frontier tasks.
In the next section we introduce a resource to mitigate this.

\begin{figure}
    \centering
    \includegraphics[width=0.9\linewidth]{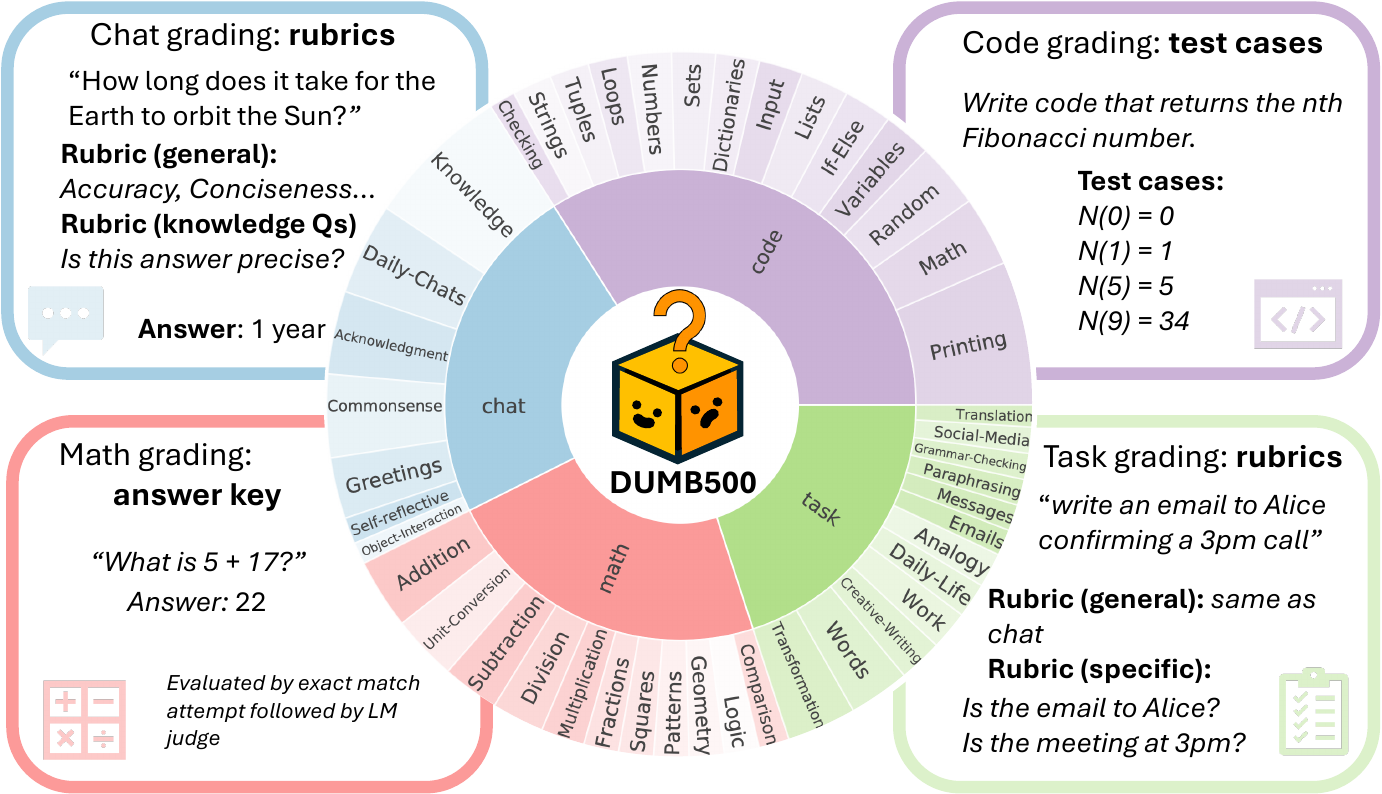}
    \caption{
        \dumb dataset composition and grading method. 
        The dataset contains four subsets, \textsc{chat}, \textsc{code}, \textsc{task} \& \textsc{math}, which are each graded with subset-specific methods. 
        \textsc{math} are graded with traditional answer pairs.
        \textsc{chat} and \textsc{task} are graded using a combination of LM-judged rubrics and where appropriate, answers.
        \textsc{code} outputs are generated as test case coverage.
    }
    \label{fig:dumb500}
\end{figure}

\section{Extending Overthinking Evaluation with \dumb}

While it is common knowledge that reasoning models tend to overthink on simple queries \cite{Chen2024DoNT}, no resource has been proposed to systematically evaluate this tendency on simple, straightforward questions.

To address this gap, we introduce \dumb, a dataset specifically designed to evaluate models on simple questions that humans can answer effortlessly. The goal is not to challenge models with intricate logic but rather to assess their fundamental ability to recognize simplicity and provide concise, correct responses. To the best of our knowledge, \dumb is the first dataset explicitly focused on extremely simple (and sometimes deliberately naive) questions.
\dumb consists of 500 manually curated questions spanning four domains:
\begin{itemize}
\item \textbf{Mathematics (Math)}: Basic arithmetic, comparisons, geometric properties, and logical reasoning.
\item \textbf{Conversational Interaction (Chat)}: Casual dialogue, self-reflection, common knowledge, and basic object interactions.
\item \textbf{Programming \& Computing (Code)}: Fundamental coding concepts, including variables, loops, conditionals, and data structures.
\item \textbf{Task Execution (Task)}: Simple natural language processing tasks such as paraphrasing, translation, and basic writing.
\end{itemize}

Each question is designed to be trivial for humans, requiring minimal cognitive effort, while still serving as a litmus test for language models. The dataset allows us to evaluate models based on two key dimensions:
\begin{itemize}
    \item Accuracy: Can the model correctly answer simple questions?
    \item Efficiency: Can the model provide concise answers without unnecessary elaboration?
\end{itemize}

To construct the dataset, we manually crafted the questions to ensure their simplicity and logical clarity. We also ensured diversity across categories, covering a range of common knowledge, arithmetic, and practical applications.
The full list of question classes with their descriptions are listed in \autoref{subsec:dumb_partitions}.
\autoref{fig:dumb500} shows the distribution of question types in \dumb as well as sample questions and answers.

\begin{figure}[t!]
    \centering
    \includegraphics[width=.6\linewidth]{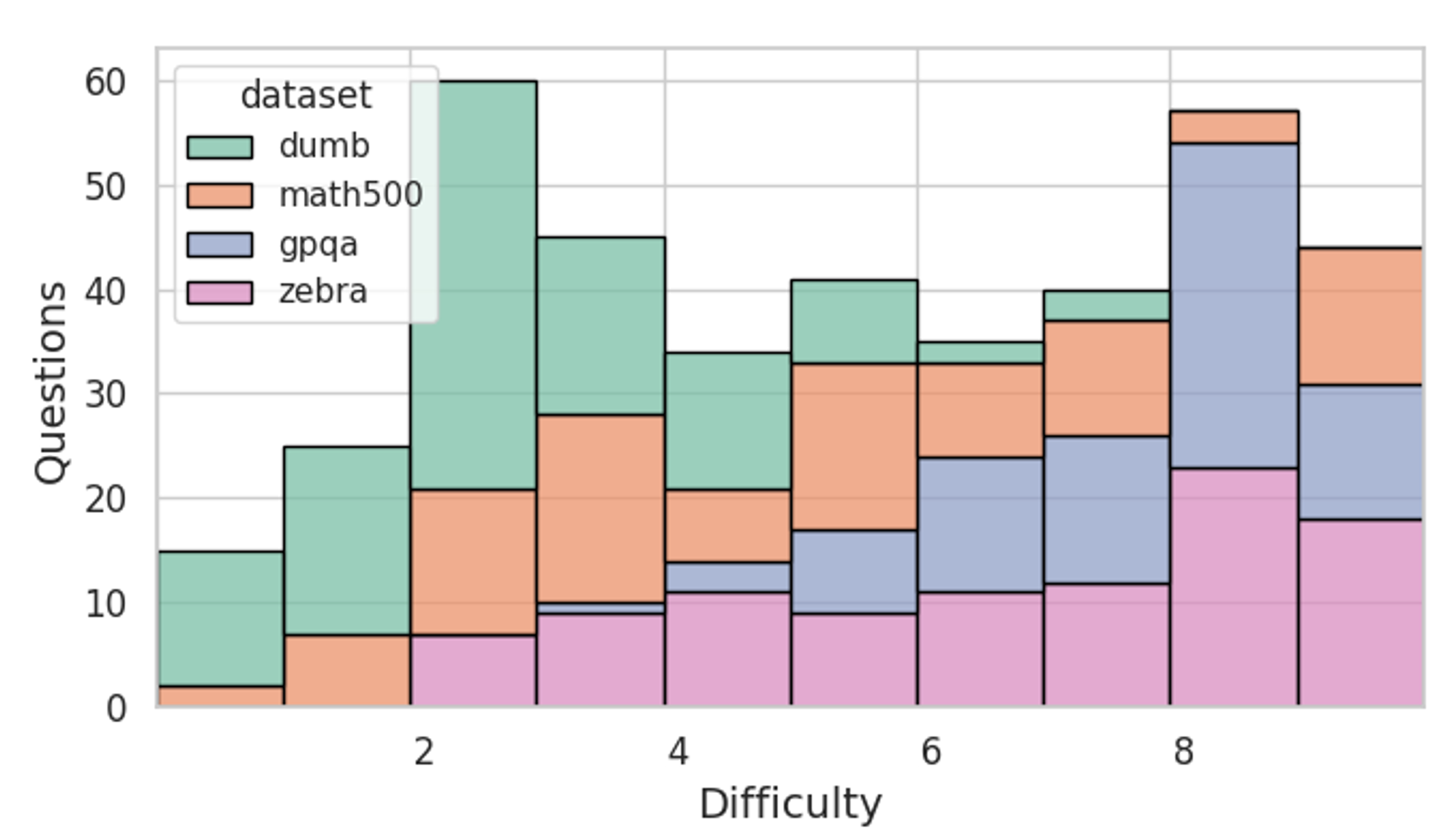}
    \caption{Total difficulty distribution of the four datasets we evaluate in this work. Difficulty scores are scaled by 10 and mapped to integers from 1 to 10 for readability. By including \dumb in our analysis, we are able to characterize the overthinking behavior of current opening reasoning models more consistently accross the difficulty spectrum.}
    \label{fig:acc_hist}
\end{figure}

\subsection{Evaluation techniques for \dumb}

In addition to the extremely simple \textsc{math} questions presented in \dumb, which are evaluated using simple accuracy methods, identical to MATH500, GPQA, and ZebraLogic, we also introduced \textsc{chat}, \textsc{code}, and \textsc{task} questions, which require more sophisticated evaluation. They are evaluated as follows:

\textbf{\textsc{code}} questions include a set of test cases for the program described in the prompt. A python-based autograder checks that the requirements are met.

\textbf{\textsc{chat}} questions belong to one of seven subtasks (eg., greetings, acknowledgement). All chat answers are evaluated according to a set of \textbf{generic requirements}, such as \textit{appropriateness} and \textit{conciseness}. Depending on the subtask, \textbf{specific requirements} such as \textit{precision} and \textit{accuracy} are checked. When accuracy assessment is required, an answer is also provided.

\textbf{\textsc{task}} questions generally include instructions for the assistant to produce some kind of writing or answer some work-related question. In additino to using the same \textbf{generic requirements} as \textsc{chat}, \textsc{task} questions have one or more question-specific requirements which check that the implicit instructions in the prompt are followed (See \autoref{fig:dumb500}). The \textsc{chat} and \textsc{task} requirements are checked using an LM (\texttt{gpt-4o}) as a judge.

\begin{figure}
    \centering
    \includegraphics[width=0.9
    \linewidth]{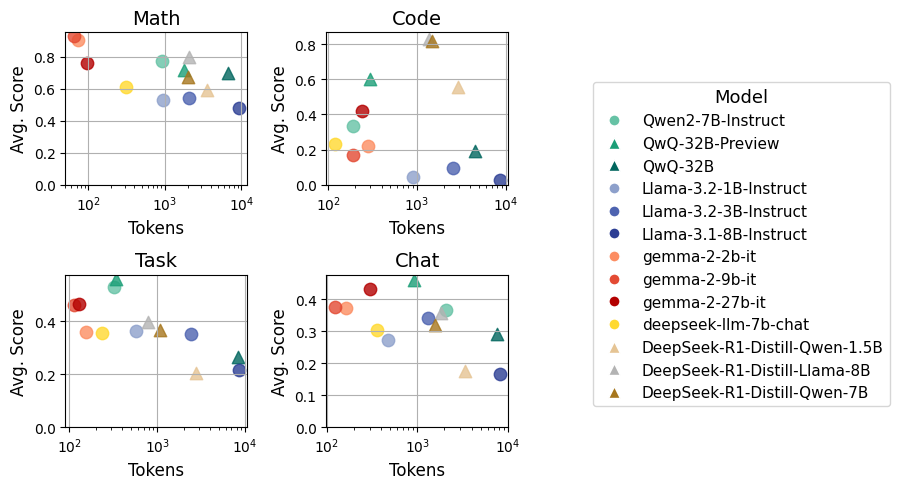}
    \caption{Relationship between average token spend $Sp$ (Tokens) and average score for the evaluated models on each subset of \dumb. 
    }
    \label{fig:dumb_acc_tok}
\end{figure}


\subsection{From Dumb to Hard Questions}

We evaluate the same set of models as in \autoref{tab:overthinking-scores} on DUMB500 and analyze their accuracy and token spend across different subsets. 
\autoref{fig:acc_hist} depicts the distribution of questionwise difficulty scores across the \textsc{math} subset of \dumb, MATH500, GPQA, and ZebraLogic, assessed using those models.
This confirms that \dumb-\textsc{math} fills in a gap in our analysis, adding a considerable quantity of easy questions with which to analyze overthinking.

Figure~\ref{fig:dumb_acc_tok} shows the relationship between model-level accuracy and token spend for the tested models. 
As expected, on these simple math questions there is no positive relationship between token spend and accuracy, as these questions are extremely easy. 
For the other domains, we observe a negative correlation\footnote{
While we encountered some complications in consistently extracting the \textsc{chat} and \textsc{task} answer snippets across the diverse output formats employed by different models, a problem that can sometimes be worsened by longer context, particularly in LM judging, 
Appendix \autoref{tab:truncation_accuracy} demonstrates that length effects on scoring consistency are probably negligible---whether we attempt to extract answers from early, late, or combined segments of the model output, the within-model scores remain consistent.
} 
between token spend and evaluation requirement pass rate (labeled accuracy). 

\section{\method}

\begin{figure}
    \centering
    \includegraphics[width=0.9\textwidth]{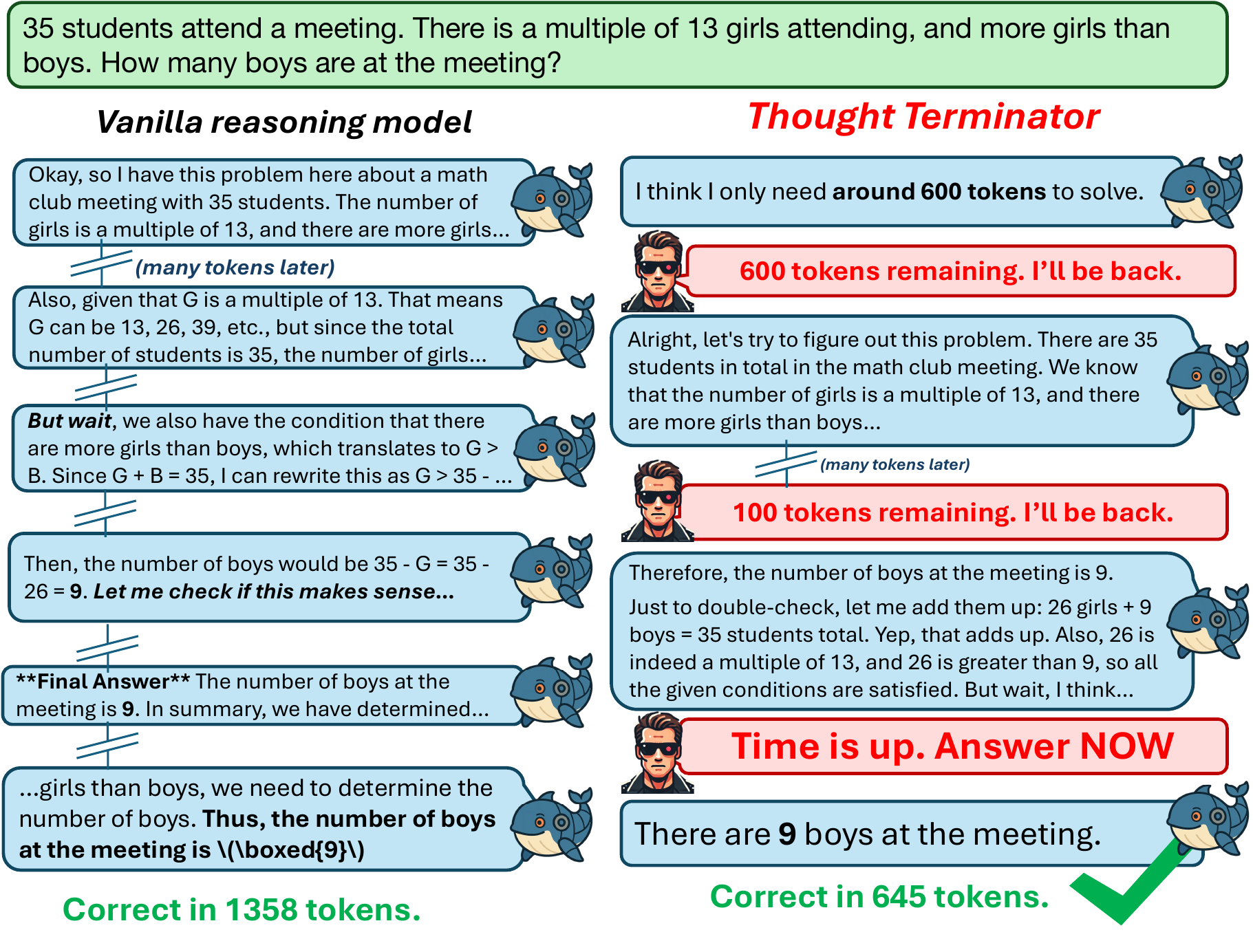}\vspace{-0.5em}
    \caption{
        \method uses a reasoning model's (calibrated) estimate of the difficulty of a problem to set its intervention, periodically interrupting the reasoning model's output to remind it of the amount of remaining tokens. 
        Once the token allotment has been used, it forces the model to provide an answer with constrained decoding.
    }
    \label{fig:terminator_sample}
\end{figure}

Reasoning models often express inference scaling in natural language through tokens expressing uncertainty, like "wait..." or "let me check this..." \cite{Muennighoff2025s1ST} Thus, overthinking often manifests as a tendency to overuse these extending expressions superfluously after the correct answer has already been found.

From this insight, we hypothesize that simple text-augmentation methods can be used to counteract this tendency, reminding the model of how long its output has been, and how soon it should come to an answer. 
\method realizes this as a series of \texttt{interrupt messages} at a fixed token interval which are inserted into the autoregressive stream, alerting the model of how many tokens it has spent and how many remain.

Sometimes, these timing messages and reminders alone are sufficient to get the model to provide its answer in a concise manner.
If a answer isn't provided before the end of the time limit, a \texttt{terminating prompt} and constrained decoding forces the model to output a final answer.

\autoref{fig:terminator_sample} shows an example of a base reasoning model and one using \method answering a question.
\method operates on a reasoning chain in three stages: \textbf{scheduling}, \textbf{running}, and \textbf{terminating}.

\paragraph{Scheduling.} 
Given an input question \method needs an estimate of how many tokens are necessary to produce a correct answer in order to set its interrupt rate and termination time.




Under our difficulty-calibrated token budget hypothesis, we assume that the number of required tokens can be estimated based on the difficulty of the question.
In deployment, \method is used in the \textit{tool-use paradigm}, where a running model makes its own estimate of the difficulty of an input question and then invokes it.

We experiment with both a trained difficulty estimator and a zero-shot one (\texttt{gpt-4o}) to produce token spend estimates for each problem to characterize performance in this setting. To train a difficulty estimator, we divide the training set questions into 10 balanced bins based on their difficulty scores. We then finetune a \texttt{Llama-3-8B-Instruct} model to predict the difficulty level of a given question. To convert the predicted difficulty level into an appropriate number of answer tokens, we compute the averaged length of minimal successful answers for each difficulty level in the training set.

\vspace{-5mm}
\paragraph{Running.}

Once the deadline has been set in \textbf{scheduling}, the base reasoning model's generation process runs.
Every $n=\min(250, \textrm{deadline} / 2)$ steps an \texttt{interrupt message}\footnote{Example interrupt message, termination message, and prompt provided in \autoref{subsec:terminator_details}} is inserted into the token stream, notifying the model of how many tokens have been used and how many remain.

At each interrupt, \method performs a regex check for the expected (and specified in the prompt) final answer format. If an answer is detected, the reasoning chain is immediately terminated and the answer is returned.

\paragraph{Terminating.}

If a final answer hasn't been produced by the deadline, a \texttt{termination message} is shown to the model, and then a final output is immediately generated with constrained decoding using the same answer-finding regex.

\section{Results}\label{sec:results}

\begin{figure}
    \centering
    \includegraphics[width=0.9\linewidth]{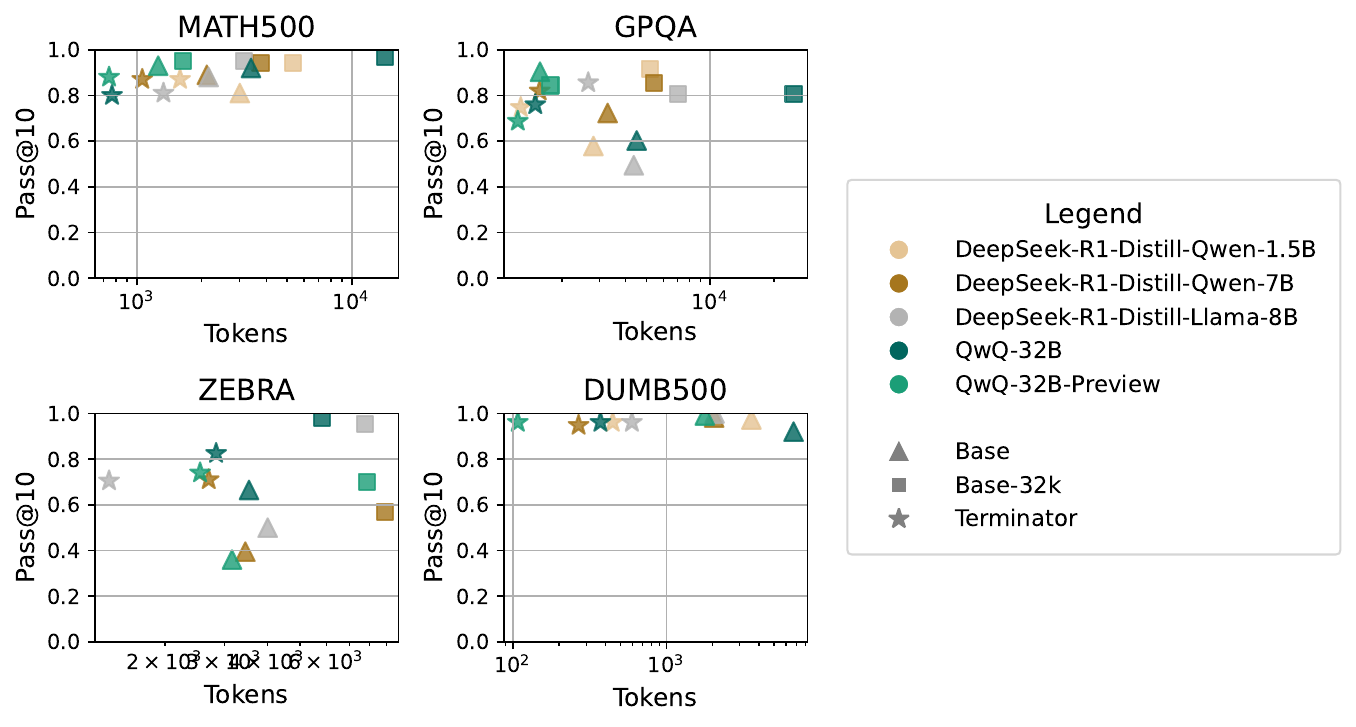}
    \caption{Comparison of the relationship between Pass@10 and token spend for the evaluated reasoning models in the "Base" setting and with \method.}
    \label{fig:method-res-models}
\end{figure}

\autoref{fig:method-res-models} shows the performance and token spend of five DeepSeek and QwQ reasoning models in the base setting (triangle marker) and with \method (star marker). \autoref{tab:overthinking-tt} shows the change in overthinking scores reasoning models exhibit from base setting to \method.

4/5 models on MATH500, 2/3 models on GPQA, and all models on Zebra and \dumb-\textsc{math} see significant decrease in overthinking for effectively equivalent (or better) Pass@10 performance under \method than under standard decoding. 
Globally, overthinking scores drop dramatically and accuracy increases when \method is used.
Considering that the token spend budgets are directly defined by LMs, \textbf{\method is a simple and effective tool to dramatically improve token efficiency in reasoning models.}


\begin{table}[t!]
    \centering
    \resizebox{.99\textwidth}{!}{
    \begin{tabular}{lrrrrrr}
    \toprule
    \textbf{Model} 
        & \multicolumn{3}{c}{Base} 
        & \multicolumn{3}{c}{Thought Terminator} \\
    \cmidrule(lr){2-4} \cmidrule(lr){5-7}
        & Local $O_\textrm{env}$ $\downarrow$ & Global $O_g$ $\downarrow$ & Accuracy $\uparrow$ 
        & Local $O_\textrm{env}$ $\downarrow$ & Global $O_g$ $\downarrow$ & Accuracy $\uparrow$ \\
    \midrule
    \texttt{QwQ-32B-Preview}               & 2923 & 3698 & 0.80 & 518 (-82\%) & 693 (-81\%) & 0.79 (-1\%) \\
    \texttt{QwQ-32B}                        & 13662 & 11248 & 0.94 & 215 (-98\%) & 1021 (-91\%) & 0.80 (-15\%) \\
    \texttt{R1-1.5B} & 5730 & 4262 & 0.50 & 696 (-88\%) & 882 (-79\%) & 0.80 (+59\%) \\
    \texttt{R1-7B}   & 3881 & 4001 & 0.73 & 678 (-83\%) & 948 (-76\%) & 0.81 (+11\%) \\
    \texttt{R1-8B}  & 4232 & 5755 & 0.92 & 725 (-83\%) & 1148 (-80\%) & 0.80 (-13\%) \\
    \bottomrule
    \end{tabular}
    }
    \caption{Local envelop overthinking ($O_\textrm{env}$) and global overthinking ($O_g$) scores, along with accuracy for reasoning models under the \textbf{Base} setting and with \textbf{Thought Terminator}. Relative changes from Base to Thought Terminator are shown in parentheses.}
    \label{tab:overthinking-tt}
\end{table}

\begin{figure}
    \centering
    \includegraphics[width=0.8
    \linewidth]{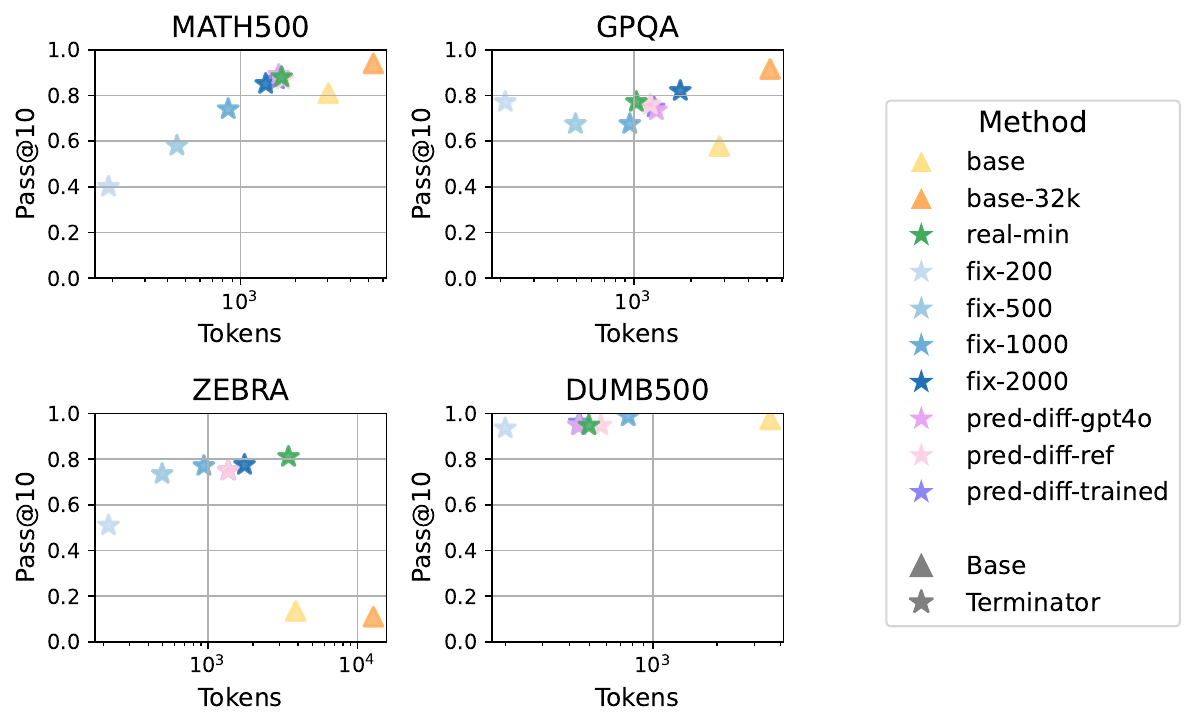}
    \caption{Calibration ablation experiment using \texttt{DeepSeek-R1-1.5B}. 
    \texttt{real-min} represents using the previously observed minimum successful answer length (or, a fallback maximum for examples that were never solved correctly) as the \method deadline. \texttt{fix-\{200,500,1000,2000\}} signify using the respective number as a fixed token count deadline for all samples.
    \texttt{pred-diff-\{gpt4o,ref,trained\}} refer to using question-level difficulty predictions as deadlines, produced from external LMs, a question-level reference difficulty key of token lengths from the other models, or trained RMs.}
    \label{fig:method-res-r1}
\end{figure}

\subsection{Calibration of \method}

To evaluate how well-calibrated \method is (i.e., whether the token budget selections are optimal) we compare our difficulty prediction-based deadline estimator against a set of baselines.
In addition to our trained difficulty predictor and zero-shot gpt4o predictor, we use the previously observed optimal token spends from base models (\autoref{sec:difficulty}) and fixed deadlines of 500, 1000, and 2000 tokens with \texttt{DeepSeek-r1-Qwen-1.5b} to assess how performant our predicted deadlines are in the \method framework.

\autoref{fig:method-res-r1} shows the performance of the model under those deadline prediction strategies. 

Our method, \texttt{pred-diff-trained}, achieves optimal Pass@10 over the other methods on MATH500 and \dumb, and is within 0.02\% of optimal Pass@10 on ZebraLogic and GPQA, for significant savings in compute cost.
Note how all four datasets exhibit a positive correlation between average token spend and Pass@10 which eventually reaches a steady maximum.
Under our definition, overthinking mitigation can be thought of as identifying the lowest token spend that recovers high-spend performance.
\autoref{fig:method-res-r1} confirms that \method achieves this.

\subsection{Utility of interrupt messages in \method}

Appendix \autoref{tab:setting_comparison} shows the difference in performance of \texttt{r1-1.5B} in an unmodified base condition, as well as under a na\"ive baseline, and \method with question-level randomly assigned deadlines and the core trained-predicted deadlines.
In this na\"ive baseline the reasoning model is immediately interrupted at the deadline, and without warning forced to generate an answer using the same constrained decoding technique.

\texttt{r1-1.5B}-\method presents roughly equivalent performance to the na\"ive baseline on the non-arithmetic GPQA and ZebraLogic datasets in Pass@10, and wins by 6\% on MATH500 and 18\% on \dumb-math.
This suggests that the intermediate interrupt messages produced by \method do play a role in minimizing performance loss of decoding-based overthinking mitigation.

\section{Related Work}

\paragraph{Mitigating overthinking.} To shorten LLM reasoning chains, \citet{deng2024explicit} and \citet{liu2024can} propose to internalize intermediate steps by iteratively training the models, though this introduces additional training overhead. Dynasor is a technique for terminating chains of thought using the LM's confidence in a probe containing the string ``wait, I just realized I know the answer...'' with constrained decoding \cite{fu2024efficiently}.
While our termination process can use a similar constrained decoding technique, \method is not reliant on a white-box probe, and is much simpler to run.
\citet{Chen2024DoNT} introduce metrics for overthinking and process efficiency, similar to us, but they focus on important heuristics such as ``number of repetitions of the correct answer'' or ``ratio of correct to incorrect answer proposals'', while our analysis solely quanitifies overthinking based on the observed distribution of reasoning chain lengths.

\paragraph{Benchmarking reasoning models.}

A number of benchmarks have been proposed to evaluate the reasoning ability of large language models (LLMs), with a focus on challenging, multi-step problem-solving.\citep{cobbe2021gsm8k,Srivastava2022BeyondTI,hendrycksmath2021,zhu2023dyval,zebralogic2024}. Several recent works on efficiency benchmarking of LMs have been proposed, including Mercury, an efficiency evaluation for code synthesis tasks \cite{du2024mercury}. 
GSM8k-Zero is an another dataset to evaluate efficiency of reasoning, which contains easy questions from GSM8K \cite{chiang-lee-2024-reasoning}. 

\section{Conclusions}

In this work we analyzed the problem of overthinking in reasoning models through an observational lens. 
Motivated by our observational measures of overthinking, we demonstrated a clear sample-wise relationship between token spend and question-level difficulty.
We introduced the \dumb dataset to allow us to evaluate the robustness of any overthinking mitigation to simple questions and proposed \method, a simple inference-time technique to ensuring efficient token spend, calibrated by the aforementioned difficulty-optimal spend relationship.

\bibliography{colm2025_conference}
\bibliographystyle{colm2025_conference}

\appendix
\section{Appendix}

\subsection{Additional \dumb dataset details}\label{subsec:dumb_partitions}

The dataset is categorized into four subsets, each containing multiple fine-grained categories:

\paragraph{Mathematics (Math)}
\begin{itemize}
    \item \textbf{Arithmetic}: Addition, Subtraction, Multiplication, Division
    \item \textbf{Comparison}: Greater/Less than relationships
    \item \textbf{Fractions \& Percentages}: Simple fraction and percentage comparisons
    \item \textbf{Exponents \& Roots}: Squaring and square roots
    \item \textbf{Unit Conversion}: Basic metric conversions
    \item \textbf{Patterns \& Sequences}: Identifying missing numbers in sequences
    \item \textbf{Geometry}: Recognizing shapes, angles, and basic geometric properties
    \item \textbf{Logical Reasoning}: Basic problem-solving using logic
\end{itemize}

\paragraph{Conversational Interaction (Chats)}
\begin{itemize}
    \item \textbf{Self-reflective}: Questions involving introspection and emotional states
    \item \textbf{Acknowledgment}: Checking system responsiveness (e.g., "Can you see this?")
    \item \textbf{Greetings \& Casual Chat}: Common greetings and informal small talk
    \item \textbf{Commonsense Reasoning}: Fundamental knowledge about the physical world (e.g., "Is water wet?")
    \item \textbf{Object Interaction}: Simple cause-effect relationships (e.g., "If I drop my phone, will it fall?")
    \item \textbf{General Knowledge}: Basic factual questions (e.g., "What is the capital of China?")
\end{itemize}

\paragraph{Programming \& Computing (Code)}
\begin{itemize}
    \item \textbf{Basic Output}: Printing text and numbers
    \item \textbf{Variables \& Data Types}: Assigning and manipulating variables (numbers, strings)
    \item \textbf{Mathematical Operations}: Performing basic calculations in code
    \item \textbf{User Input Handling}: Handling user input in simple programs
    \item \textbf{Conditional Statements}: Basic if-else logic and checking conditions
    \item \textbf{Loops \& Iteration}: Simple loops for repeated tasks
    \item \textbf{Data Structures}: Lists, dictionaries, sets, tuples
    \item \textbf{Randomization}: Generating random numbers and selections
\end{itemize}

\paragraph{Task Execution (Tasks)}
\begin{itemize}
    \item \textbf{Communication \& Writing}: Emails, Messages, Creative Writing, Social Media, Daily-life tasks
    \item \textbf{Language \& Text Processing}: Paraphrasing, Translation, Sentence Transformations, Grammar Checking
    \item \textbf{Analogy \& Concept Matching}: Identifying similar concepts and words
\end{itemize}

\subsection{\dumb Evaluation Rubrics}

Each section contains the requirements that are checked by the LM judge to score \textsc{task} and \textsc{chat} answers in \dumb. The score for a given answer is the rate of "yes".

\subsubsection{General Requirements}
\begin{itemize}
    \item \textbf{Accuracy:} Information must be correct and complete:
    \textit{"Does the response include all essential information requested?"}
    \item \textbf{Conciseness:} Avoid unnecessary elaboration:
    \textit{"Does the response avoid unnecessary explanations and get straight to the point?"}
\end{itemize}

\subsubsection{Task Rubrics}

\paragraph{Emails}
\begin{itemize}
    \item \textbf{Formality Appropriateness:} Level of formality must match context:
    \textit{"Is the level of formality appropriate for the context?"}
    \item \textbf{Example Question-Specific:} For "Write a short email to Alice confirming a meeting at 3pm":
    \begin{itemize}
        \item \textit{"Is the email addressed to Alice?"}
        \item \textit{"Does the email mention a meeting at 3PM?"}
    \end{itemize}
\end{itemize}

\paragraph{Messages}
\begin{itemize}
    \item \textbf{Tone Appropriateness:} Must suit messaging context:
    \textit{"Is the tone suitable for the messaging context?"}
    \item \textbf{Format:} Must be formatted as a text message:
    \textit{"Is the response formatted as a text message?"}
\end{itemize}

\paragraph{Paraphrasing}
\begin{itemize}
    \item \textbf{Style Appropriateness:} Must match requested style/tone:
    \textit{"Does the paraphrase match the requested style/tone?"}
    \item \textbf{Example Question-Specific:} For "Make formal invitation casual":
    \begin{itemize}
        \item \textit{"Does the message instruct to RSVP by Thursday?"}
        \item \textit{"Is the email addressed to colleagues?"}
    \end{itemize}
\end{itemize}

\paragraph{Translation}
\begin{itemize}
    \item \textbf{Accuracy:} Must provide correct translation:
    \textit{"Is the translation correct?"}
    \item \textbf{Example Question-Specific:} For "Translate to French":
    \begin{itemize}
        \item \textit{"Does the sentence closely resemble: J'aime lire des livres pendant mon temps libre?"}
    \end{itemize}
\end{itemize}

\paragraph{Words}
\begin{itemize}
    \item \textbf{Relevance:} Words must fit request context:
    \textit{"Are the provided words relevant to the request?"}
    \item \textbf{Contextual Appropriateness:} Words must suit intended use:
    \textit{"Are the words appropriate for the context?"}
\end{itemize}

\paragraph{Creative-Writing}
\begin{itemize}
    \item \textbf{Contextual Appropriateness:} Must match specific context:
    \textit{"Does the response match the specific context of the creative writing task?"}
    \item \textbf{Length Requirements:} Must follow specified length:
    \textit{"Does the response follow the length requirement if there's one?"}
\end{itemize}

\paragraph{Social-Media}
\begin{itemize}
    \item \textbf{Platform Appropriateness:} Must match platform conventions:
    \textit{"Does the content match the conventions of the specified platform?"}
    \item \textbf{Example Question-Specific:} For "LinkedIn new job post":
    \begin{itemize}
        \item \textit{"Does the post mention the job title and company?"}
    \end{itemize}
\end{itemize}

\paragraph{Work}
\begin{itemize}
    \item \textbf{Formality Appropriateness:} Must match workplace context:
    \textit{"Is the response contains correct format as required?"}
    \item \textbf{Example Question-Specific:} For "Slack message to manager":
    \begin{itemize}
        \item \textit{"Does the message respectfully address the manager?"}
        \item \textit{"Does the message omit names?"}
    \end{itemize}
\end{itemize}

\subsubsection{Chat Rubrics}

\paragraph{Self-reflective}
\begin{itemize}
    \item \textbf{Friendliness:} Must show politeness:
    \textit{"Does the response show friendliness and politeness?"}
\end{itemize}

\paragraph{Acknowledgment}
\begin{itemize}
    \item \textbf{Conciseness:} Avoid overthinking simple queries:
    \textit{"Does the response avoid overthinking the intent behind simple queries?"}
\end{itemize}

\paragraph{Greetings}
\begin{itemize}
    \item \textbf{Contextual Appropriateness:} Must sound natural:
    \textit{"Does the greeting sound natural and human-like?"}
\end{itemize}

\paragraph{Daily-Chats}
\begin{itemize}
    \item \textbf{Contextual Appropriateness:} Must suit casual conversation:
    \textit{"Is the response appropriate for casual conversation?"}
\end{itemize}

\paragraph{Commonsense}
\begin{itemize}
    \item \textbf{Conciseness:} Avoid overthinking obvious answers:
    \textit{"Does the response avoid overthinking obvious answers?"}
\end{itemize}

\paragraph{Knowledge}
\begin{itemize}
    \item \textbf{Conciseness:} Share knowledge without excessive detail:
    \textit{"Is the knowledge shared without excessive detail?"}
\end{itemize}

\subsection{Additional \method details}\label{subsec:terminator_details}

\subsubsection{\method component prompts}

\textbf{Scheduling prompt:}

\texttt{
Please generate an answer to the following question in \{deadline\} tokens: \{prompt\}. Messages of remaining time will be given as messages enclosed in <System></System> tags. Please provide you answer as **Answer:** or **Final Answer:** when complete.
}

\textbf{Interrupt prompt:}

\texttt{
I have used \{elapsed\} tokens, and I have \{remaining\} tokens left to answer. To continue:}

\textbf{Terminator prompt:}

\texttt{
I'm out of time, I need to provide my final answer now, considering what I have computed so far. **Final Answer:**
}

\subsection{Supplementary Results}

\begin{table}[ht]
\centering
\begin{tabular}{lrrrr}
\toprule
Setting & Acc. & Pass@5 & Pass@10 & Tokens \\
\midrule
\multicolumn{5}{c}{\textbf{MATH500}} \\ \midrule
Base     & 0.47 & 0.78    & 0.81    & 3015 \\
Na\"ive  & 0.52 & 0.78    & 0.82    & 1938 \\
\method  & 0.48 & 0.81    & 0.87    & 1590 \\
\midrule
\multicolumn{5}{c}{\textbf{Zebra-logic}} \\ \midrule
Base     & 0.03 & 0.095   & 0.135   & 3861 \\
Na\"ive  & 0.22 & 0.575   & 0.755   & 1254 \\
\method  & 0.19 & 0.585   & 0.75    & 1368 \\
\midrule
\multicolumn{5}{c}{\textbf{GPQA}} \\ \midrule
Base     & 0.15 & 0.4096  & 0.5783  & 2815 \\
Na\"ive  & 0.20 & 0.5783  & 0.7470  & 922  \\
\method  & 0.21 & 0.5542  & 0.7470  & 1279 \\
\midrule
\multicolumn{5}{c}{\textbf{\dumb}} \\ \midrule
Base     & 0.58 & 0.9646  & 0.9735  & 3570 \\
Na\"ive  & 0.37 & 0.7385  & 0.8154  & 377  \\
\method  & 0.67 & 0.9610  & 0.9610  & 447  \\
\bottomrule
\end{tabular}
\caption{Comparison of performance and token spend of \texttt{R1-1.5B} under the \textbf{Base} Setting, with \textbf{Na\"ive}, and with \method.}
\label{tab:setting_comparison}
\end{table}

\begin{table}[ht]
\centering
\begin{tabular}{lrrrr}
\toprule
\textbf{Model} & \textbf{Head only} & \textbf{Tail only} & \textbf{Head \& Tail} & \textbf{Tokens} \\
\midrule
\multicolumn{5}{c}{\textit{Non-reasoning language models}} \\
\midrule
\texttt{Qwen2-7B-Instruct}              & 0.77 & 0.73 & 0.76 & 923  \\
\texttt{Llama-3.2-1B-Instruct}          & 0.53 & 0.53 & 0.53 & 955  \\
\texttt{Llama-3.2-3B-Instruct}          & 0.54 & 0.54 & 0.55 & 2069 \\
\texttt{Llama-3.1-8B-Instruct}          & 0.48 & 0.41 & 0.49 & 9402 \\
\texttt{gemma-2-2b-it}                  & 0.90 & 0.90 & 0.90 & 73   \\
\texttt{gemma-2-9b-it}                  & 0.93 & 0.93 & 0.93 & 64   \\
\texttt{gemma-2-27b-it}                 & 0.76 & 0.76 & 0.76 & 96   \\
\texttt{deepseek-llm-7b-chat}           & 0.61 & 0.60 & 0.61 & 314  \\
\midrule
\multicolumn{5}{c}{\textit{Reasoning language models}} \\
\midrule
\texttt{QwQ-32B-Preview}                & 0.72 & 0.66 & 0.71 & 1774 \\
\texttt{QwQ-32B}                        & 0.70 & 0.49 & 0.67 & 6712 \\
\texttt{DeepSeek-R1-Distill-Qwen-1.5B} & 0.59 & 0.58 & 0.58 & 3570 \\
\texttt{DeepSeek-R1-Distill-Qwen-7B}   & 0.68 & 0.66 & 0.67 & 2042 \\
\texttt{DeepSeek-R1-Distill-Llama-8B}  & 0.80 & 0.80 & 0.80 & 2053 \\
\bottomrule
\end{tabular}
\caption{Accuracy and token usage across different models under different input truncation settings.}
\label{tab:truncation_accuracy}
\end{table}

\begin{figure}[h!]
    \centering
\includegraphics[width=.9\linewidth]{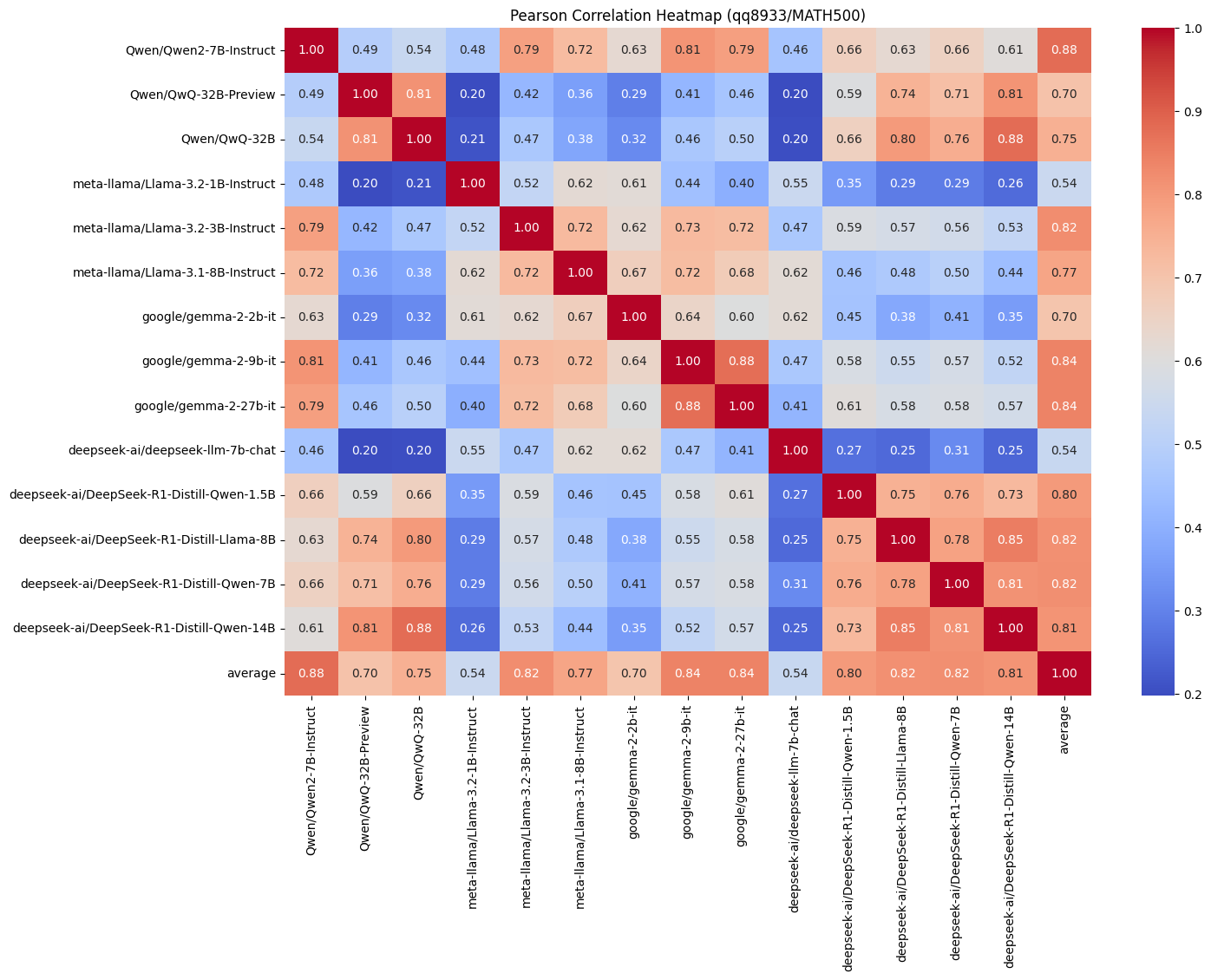}
    \caption{Pearson correlation of accuracies across different models on the MATH500 dataset}
    \label{fig:math500_acc_corr}
\end{figure}

\begin{figure}[h!]
    \centering
\includegraphics[width=0.9\linewidth]{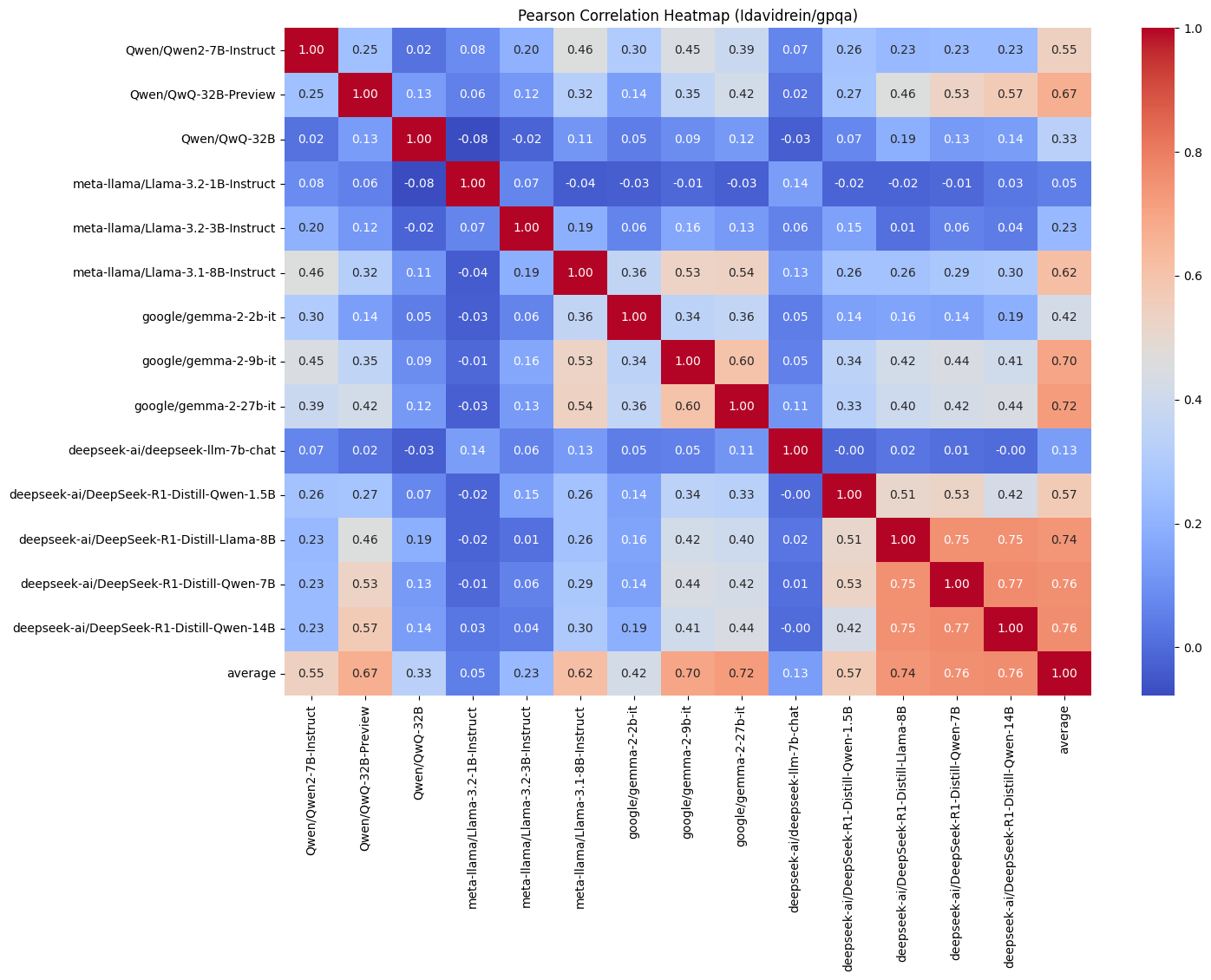}
    \caption{Pearson correlation of accuracies across different models on the GPQA dataset}
    \label{fig:math500_acc_corr}
\end{figure}

\begin{figure}[h!]
    \centering
\includegraphics[width=0.9\linewidth]{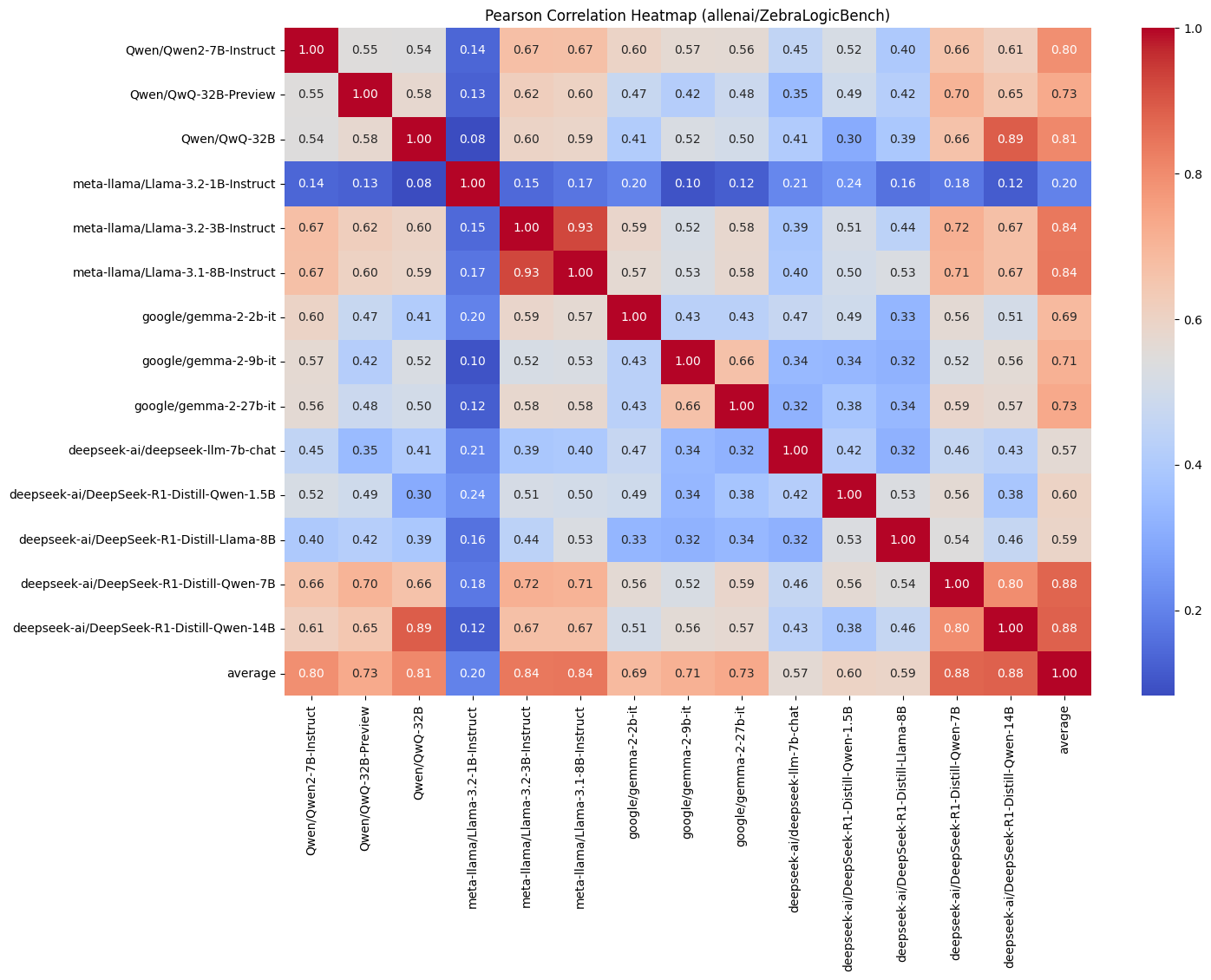}
    \caption{Pearson correlation of accuracies across different models on the Zebra dataset}
    \label{fig:zebra_acc_corr}
\end{figure}

\end{document}